\documentstyle[12pt,psfig]{article}

\def\sspace{\multiply\normalbaselineskip 200
                 \divide\normalbaselineskip 300 \normalbaselines
                 \csname @@normalbaselineskip\endcsname\normalbaselineskip}
\def\dspace{\multiply\normalbaselineskip 150
                  \divide\normalbaselineskip 100 \normalbaselines
                  \csname @@normalbaselineskip\endcsname\normalbaselineskip}

\begin{document}

\title{A Human - machine interface for teleoperation of \\
arm manipulators in a complex environment}

\author{I. Ivanisevic and V. Lumelsky\\
University of Wisconsin-Madison\\
Madison, Wisconsin 53706, USA\\}
\date{}

\maketitle

\begin{abstract}
This paper discusses the feasibility of using {\em configuration
space} (C-space) as a means of visualization and control in
operator-guided real-time motion of a robot arm manipulator.  The
motivation is to improve performance of the human operator in tasks
involving the manipulator motion in an environment with
obstacles. Unlike some other motion planning tasks, operators are
known to make expensive mistakes in such tasks, even in a simpler
two-dimensional case. They have difficulty learning better
procedures and their performance improves very little with
practice. Using an example of a two-dimensional arm manipulator, we
show that translating the problem into C-space improves the operator
performance rather remarkably, on the order of magnitude compared to
the usual work space control. An interface that makes the transfer
possible is described, and an example of its use in a virtual
environment is shown.
\end{abstract}

\section{Introduction}

The goal in this project is to improve the performance of human
operators in tasks that involve motion planning and control of
complex objects in environments with obstacles. The human
performance in such tasks is known to be patently inferior. Our
focus is on developing a visual computer interface that would allow
the operator to visualize and perform the work in the task {\em
configuration space} (C-space) rather than in the {\em work space}
(W-space) as usually done. To make it feasible, a computer
intelligence is provided that works alongside with human
intelligence in real time. To this effect, we combine the
``desirable'' features of human and machine intelligence and exploit
their individual strengths. This area belongs to the field of
human-centered systems, which has seen growing interest in recent
years. The intent of this work is to be applicable to many existing
research \cite{Lin} and commercial problems \cite{Schenker, Hunter}.

There is a large and rapidly developing class of technical systems
that are dependent on human contribution for their operation. In
various teleoperated systems (such as in space, nuclear reactors,
chemical cleanup sites, underwater probes) human operators plan and
guide the motion of remotely situated devices through interaction
with computer displays or three-dimensional models of the
device. Familiar examples include control of the NASA Shuttle arm
and of the Titanic exploration probe. In such tasks operators are
known to make mistakes of overlooking collisions with surrounding
objects; this results in expensive repairs and limits the system
effectiveness. People seem to be unable to navigate and manipulate
remote equipment without colliding with objects in the environment.

Similar problems occur in other settings. Guiding the position of a
robotic welding gun or spray painting device with a simultaneous
translation and orientation adjustment seems to be particularly
difficult for people, even when visual feedback is provided. 
Performance is very poor in a variety of these movement
planning tasks when time is not a constraint (the Shuttle arm, for
example); it becomes progressively worse in real-time operation, in
three-dimensional (3D) vs 2D tasks, and when system dynamics are
involved (masses, inertia etc.). (Underwater exploration probes, for
example, cannot stop while the operator considers the next move).

Experiments with human subjects \cite{Liu:a, Liu:b} suggest that 
the problem is in the peculiarities of human spatial reasoning: humans have
difficulty handling simultaneous interaction with objects at
multiple points of the device's body, or motion that involves
mechanical joints (such as in arm manipulators), or dynamic tasks.
Learning and practice improve the performance rather little. 
Furthermore, the performance pattern is the same when operating 
a physical rig or performing the task on a computer screen and 
moving the arm links with a mouse (see more on this in Section 
~\ref{sec:Results}).

\begin{figure}[t]
\centerline{
\begin{tabular}{c}
\psfig{figure=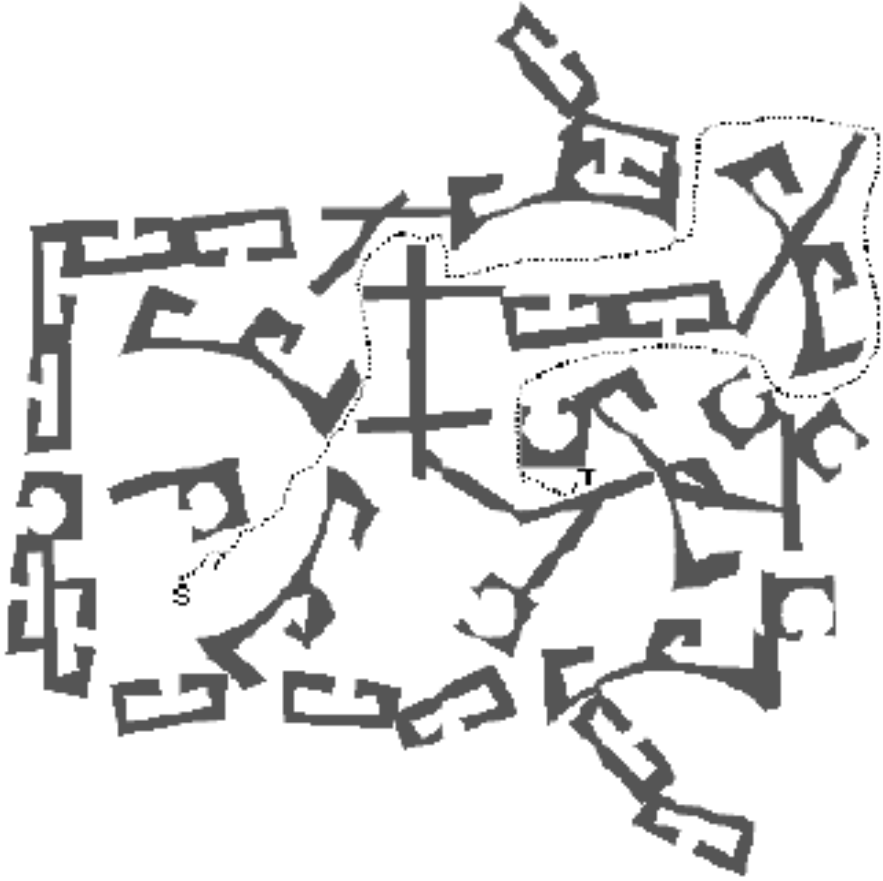,width=5in} 
\end{tabular}
} 
%\figurecaption{ 
\caption{\small Human performance in a maze. }
\label{maze}
\end{figure}

On the other hand, these experiments confirm the expected fact that
in a maze-searching problem, if information is provided about the
whole maze (a bird's-eye view), human performance is well above the
fastest computer with the best known algorithms \cite{Lum}. Figure ~\ref{maze}
gives an example of human performance in a maze: after inspecting the 
maze for a few seconds, the subjects grasp the problem and produce an
almost optimal path from point S to point T.
This contrast in the subjects' performance in the two tasks above
poses a question as to whether a human-machine interface, perhaps
with adequate machine intelligence, can be developed to improve
human performance is such applications. The current work is
an attempt to answer this question. The system we chose to model the
problem is a two-dimensional (2D) revolute-revolute (RR) arm
manipulator operating in an environment with unknown stationary
obstacles (see Figure ~\ref{arm}). The arm has two links moving in a plane,
and two revolute joints (degrees of freedom). The idea it to present
the problem to the human as one of moving a point in a maze (a task
that humans are good at) rather than the actual problem of moving a
jointed kinematic structure (which humans are not good at). We
exploit the fact that for today's computer algorithms, which are
based on spatial geometry and topology tools, both tasks present
essentially the same maze-searching problem \cite{Skewis}. By transforming the
problem to the arm {\em configuration space} (C-space), the arm is
shrunk to a point in the space of its control variables.

Below, the properties of work space control are discussed in Section
~\ref{sec:Wspace}, and those of the configuration space -- in Section 
~\ref{sec:Cspace}.  The proposed interface is then presented in Section 
~\ref{sec:Interface}, followed by some experimental results in Section 
~\ref{sec:Results} and discussion in Section ~\ref{sec:Discussion}.

\section{Work Space Control}
\label{sec:Wspace}

\begin{figure}[t]
\centerline{
\begin{tabular}{c}
\psfig{figure=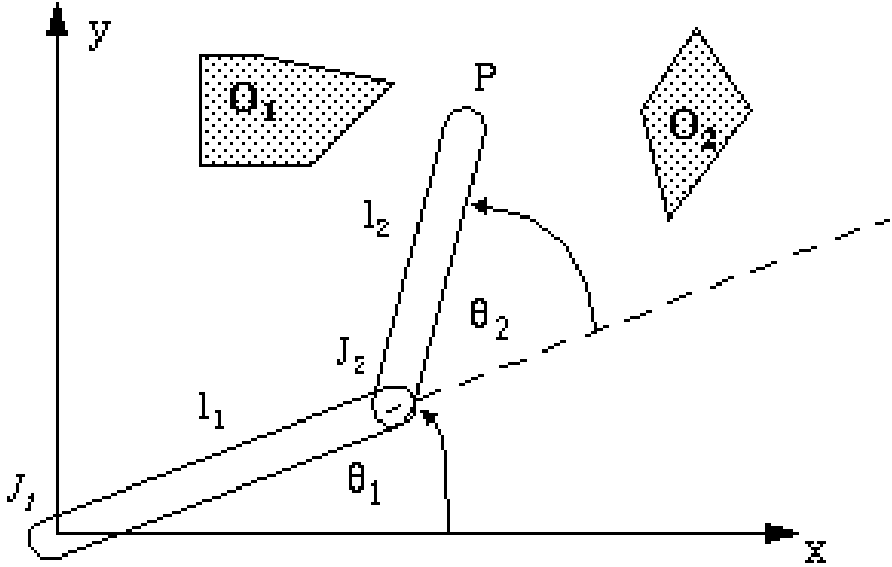,width=5in} 
\end{tabular}
} 
%\figurecaption{ 
\caption{\small The 2D two-link arm manipulator in an environment 
with unknown stationary obstacles. }
\label{arm}
\end{figure}

\begin{figure}[t]
\centerline{
\begin{tabular}{c}
\psfig{figure=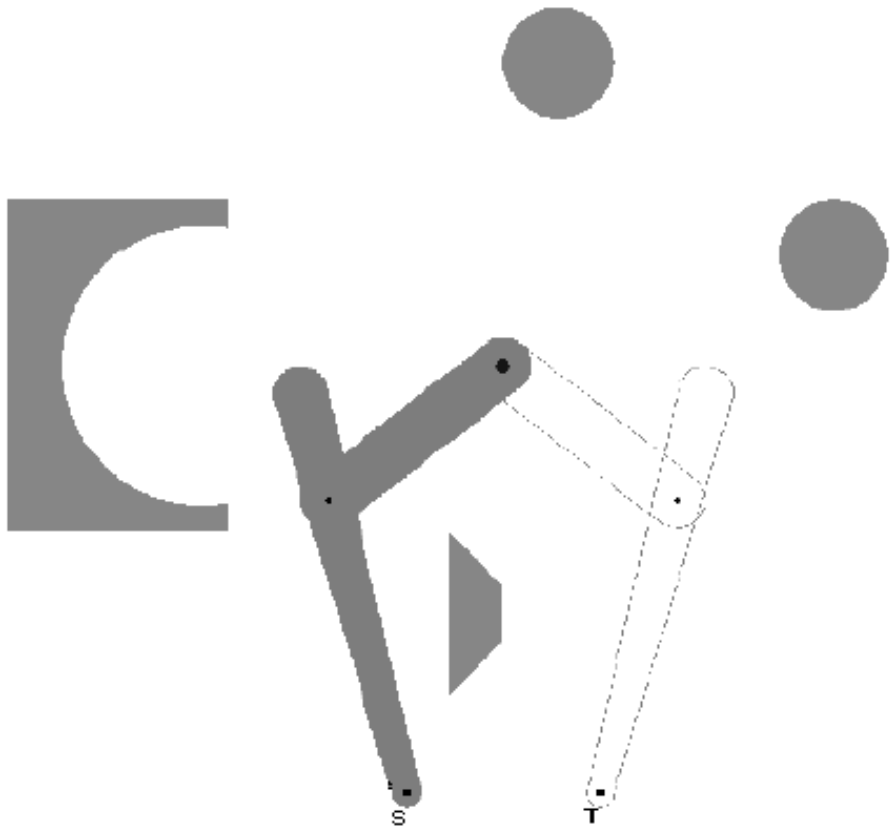,width=5in} 
\end{tabular}
} 
%\figurecaption{ 
\caption{\small A sample task. }
\label{task}
\end{figure}

The revolute-revolute (RR) planar arm considered is as follows,
Figure ~\ref{arm}: Joint $J_1$ (the shoulder) is attached to the floor, and is
the origin of a fixed reference system. Joint $J_2$ (the elbow)
connects the two links, $l_1$ and $l_2$. The Cartesian coordinates of
the endpoint (point P) are $(x, y)$. 
Moving the arm involves changing the joint angles $\theta_1$ and $\theta_2$. 
There are fixed obstacles in the arm environment ($O_1$ and $O_2$, Figure 
~\ref{arm}). There are no constraints on the shape of the obstacles 
or the arm links. The task is to move the arm from a position S 
(Start) to the position T (Target), Figure ~\ref{task}.

\subsection{Motion Control in W-space}

Arm motion is controlled with the computer mouse, in two separate
modes - joint-mode and tip-mode \cite{Skewis}. The former allows control of
individual joints by positioning the pointer closer to one of the
joints and pressing the left mouse button, which causes the selected
joint to move and align with the pointer. The tip-mode allows
control of the endpoint through the use of the middle mouse button;
computer software then calculates the corresponding joint angles of
the links and positions them accordingly.

\begin{figure}[t]
\centerline{
\begin{tabular}{c}
\psfig{figure=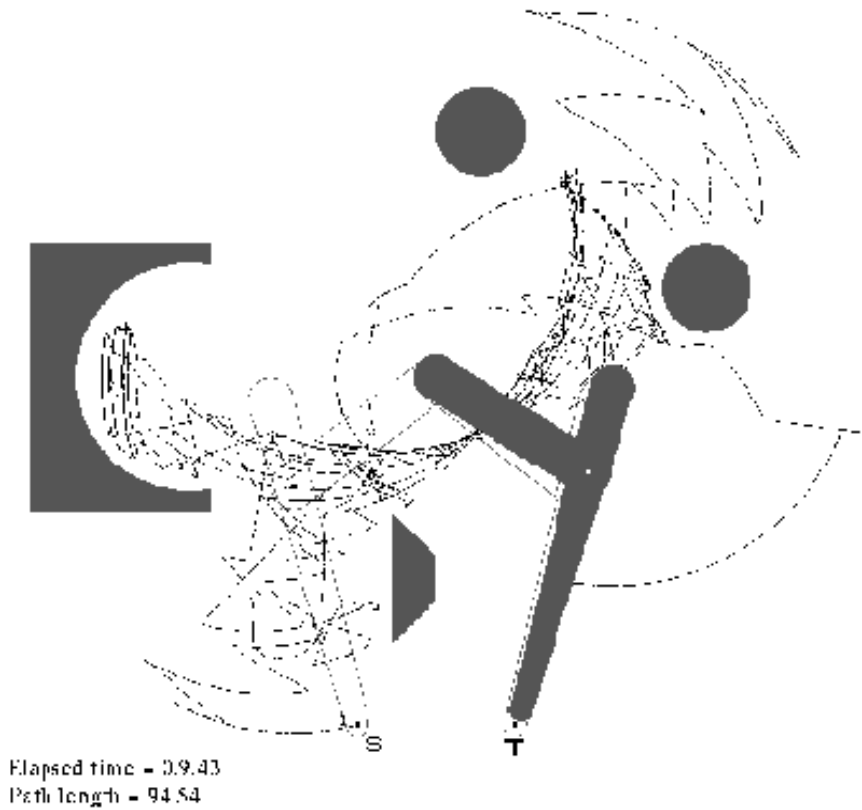,width=5in} 
\end{tabular}
} 
%\figurecaption{ 
\caption{\small An example of average human performance in W-space motion 
control. }
\label{examp1}
\end{figure}

In the joint-mode, the algorithm computes a unit vector which
describes the straight-line direction from current configuration to
specified target configuration. Assuming the configuration does not
violate step constraints (if the distance to it is larger than the
configured step size, a new target is computed by multiplying the
direction vector by step size), the new configuration becomes the
specified configuration. In tip-mode, the direction vector describes
the new position of the arm endpoint (again, subject to step
constraints), and so one needs to recover the new arm configuration
from the endpoint position $(x, y)$. This is done via the inverse
kinematics equations:
\begin{eqnarray}
\theta_2 &=& \arccos(\frac{x^2 + y^2 - l_1^2 - l_2^2}{2l_1l_2}) \label{eq1} \\
\theta_1 &=&\arctan(\frac{y}{x}) - \arctan(\frac{l_2\sin\theta_2}
{l_1 + l_2\cos\theta_2}) \label{eq2}
\end{eqnarray}
The current arm configuration is used to resolve multiple solutions
that are given by the inverse kinematics. The final step in either
motion mode is to determine if the new configuration would place the
arm in contact with the obstacle and, if so, disallow the movement
and wait for further operator input. Figure ~\ref{examp1} shows an example of
average human performance in W-space motion control; the dotted line
is the trajectory of the arm endpoint along the way from S to T. The
path length is the integral of changes in both joint angles along
the way.

\subsection{Characteristics of the Work Space Control}

Aside from being the traditional method used, W-space control has
some desirable properties:

\begin{itemize}
\item Interaction with the physical arm and its environment makes it
easier for the operator to visualize the global navigation, such as
to determine the next target configuration based on some scene
property; e.g. the operator may decide to move the arm such that its
left side will be in proximity of some object.

\item If the obstacles layout is not of much constraint on the arm motion,
this approach can yield very good (near optimal in terms of path length 
and time taken) results.

\item Given the familiar physical layout, it may be easier for the
operator to benefit from memorization of motion and improve with
training.
\end{itemize}

However, this type of control also has some serious drawbacks which
may outweigh its positive sides:

\begin{itemize}
\item In tip-mode, calculating the inverse kinematics becomes
progressively more complex and time-consuming as the number of
joints increases.

\item In a complex environment, the operator may have hard time
determining which direction of local motion is better, or whether a
given direction leads to a ``dead-end''. This is a serious drawback:
for example, in Figure ~\ref{examp1} one can pass obstacle $O_3$ with the 
elbow to the left or to the right; one of those turns out to be wrong as it
leades to a dead end, and this would become clear only significantly
later.  

\item From the standpoint of motion planning, a complex environment is not
necessarily one with many or with large obstacles; this is much
clearer in C-space (see Section ~\ref{sec:Cspace}) than in W-space.
\end{itemize}

Consequently, W-space control is likely to produce redundant motion:
as illustrated in Figure ~\ref{examp1}, the operator will often try, backtrack,
try again, backtrack again, and so on until the passage is found,
not rarely through blind luck. This also endangers the arm, as all
such motion multiplies potential collisions with surrounding
objects.  While most people do benefit from a training period in
such systems, the training can be costly (in terms of equipment use
and damage inflicted on the arm) and time consuming.

\section{Configuration Space Control}
\label{sec:Cspace}

The arm is the same 2D RR arm manipulator described in Section ~\ref{sec:Wspace}
(Figure ~\ref{arm}). Assume that the arm is capable of gathering information
about the objects in its environment via its sensors.  To simplify
the discussion, assume that those are tactile sensors - i.e. the arm
can detect an obstacle when it comes in contact with one.  The human
operator can of course view the entire workspace, Figure ~\ref{task}.  The
task is as before - to move the arm from position S to position T in
the arm's work space.

The arm can be defined in terms of the shoulder angle $\theta_1$ and the
elbow angle $\theta_2$. The set of all configurations ($\theta_1$, $\theta_2$) 
define the arm's {\em configuration space} (C-space), which can be represented 
as the surface of a common two-torus. An arm configuration in W-space
corresponds to a point in C-space. This mapping preserves
continuity; small change in W-space position corresponds to a small
change in the C-space position. A geodesic line between two points
on the torus (a straight line in the plane ($\theta_1$, $\theta_2$)) is the
``shortest path'' between the points: four such paths can actually
appear \cite{Lum}.

\subsection{Motion Control in C-space}

We will now attempt to control the arm motion indirectly, via its
point image in C-space (C-point). Each time the operator moves the
C-point slightly, the algorithm recovers a new set of configuration
variables ($\theta_1$, $\theta_2$) from the C-point coordinates and 
automatically translates it into the actual motion in W-space. That is, 
after the direction vector is calculated and step size is taken into account,
similar to the joint-mode in W-space control, angles $\theta_1$ and $\theta_2$
become available, and they are used to control the arm's next step.
Though not necessary for control purposes, for convenience a W-space
window with the arm real-time motion is shown next to the C-space
window used by the operator.

Recall that the Cartesian position $(x, y)$ of the arm endpoint is the
tip-mode parameter in W-space control. Certain applications,
e.g. grasping, may require knowledge of this parameter. If
necessary, $(x, y)$ values can be recovered from C-space information
via the direct kinematics equations:
\begin{eqnarray}
x &=& l_1\cos\theta_1 + l_2\cos(\theta_1 + \theta_2) \label{eq3} \\
y &=& l_1\sin\theta_1 + l_2\sin(\theta_1 + \theta_2) \label{eq4}
\end{eqnarray}
We are now one step away from converting the complex problem of
W-space control to a simpler problem of navigating a point in the
maze (C-space). What is missing is the maze itself. This is done by
computing the {\em C-space obstacles}, also called {\em virtual
obstacles}. Each point of a virtual obstacle corresponds to an arm
configuration that is not attainable because of interference with
the corresponding physical obstacle. The related $(x, y)$ positions in
W-space may or may not be occupied by an obstacle - in the latter
case such pieces of an obstacle are called its {\em shadows}. A
finite number of obstacles in W-space produce a finite set of
virtual obstacles in C-space.  The boundaries of virtual obstacles
are known to consist of simple closed curves \cite{Lum}. Since virtual
obstacles are defined in terms of arm variables ($\theta_1$, $\theta_2$), their
shape is visually unrelated to the shape of the W-space obstacles
\cite{Skewis, Lozano}.

\subsection{Construction of C-space Obstacles}

The greatest improvement in the operator performance comes when full
information (the bird's-eye view) about C-space is available (on the
issue of operating with uncertainty, see the discussion in Section
~\ref{sec:Discussion}).  We thus need to compute and display all the virtual
obstacles. An intuitive approach proposed in \cite{Lozano} is to treat each
link separately. First, link $l2$ is ignored, and all free space points
for the whole range of values $\theta_1$ are computed. Points of the arm
contact with physical (W-space) obstacles are recorded and become
part of the corresponding virtual obstacle. Then, for each value of
$\theta_1$ within its appropriately digitized range, free space points
for the whole range of values $\theta_2$ are computed in the same fashion,
by rotating link $l2$ around positions of joint 2 determined by the
current value $\theta_1$. Depending on the representation chosen for the
virtual obstacles, their ``insides'' may have to be filled using a
polygon-filling algorithm.

For the two-dimensional problem in question, C-obstacle calculation
can be greatly simplified by tracing the obstacle boundaries with
arm links and thus immediately creating their C-space images. Note
that two or more W-space obstacles can produce a single virtual
obstacle; that is, for the purposes of motion control, they would
indeed be one obstacle. Our simulation uses an efficient variation
of this procedure \cite{Skewis}, which makes use of the Bug1 \cite{Lum} 
algorithm.

In brief, procedure Bug1 operates as follows: the point robot starts
moving along the straight line towards target point T. If it
encounters an obstacle, a hit point H is defined at the encounter
location, and the robot turns and moves in a prespecified direction
(say, left) along the obstacle boundary.  Once the obstacle is
circled, and H is once again encountered, a leave point L is defined
on the obstacle, which is the point closest to T. The robot then
takes the shortest path to L and then proceeds to T in straight-line
fashion, repeating the procedure whenever an obstacle is
encountered. The procedure converges to T or informs that this is
impossible if true. The algorithm's computational complexity is
linear in the perimeters of the W-obstacles. 

\begin{figure}[t]
\centerline{
\begin{tabular}{c}
\psfig{figure=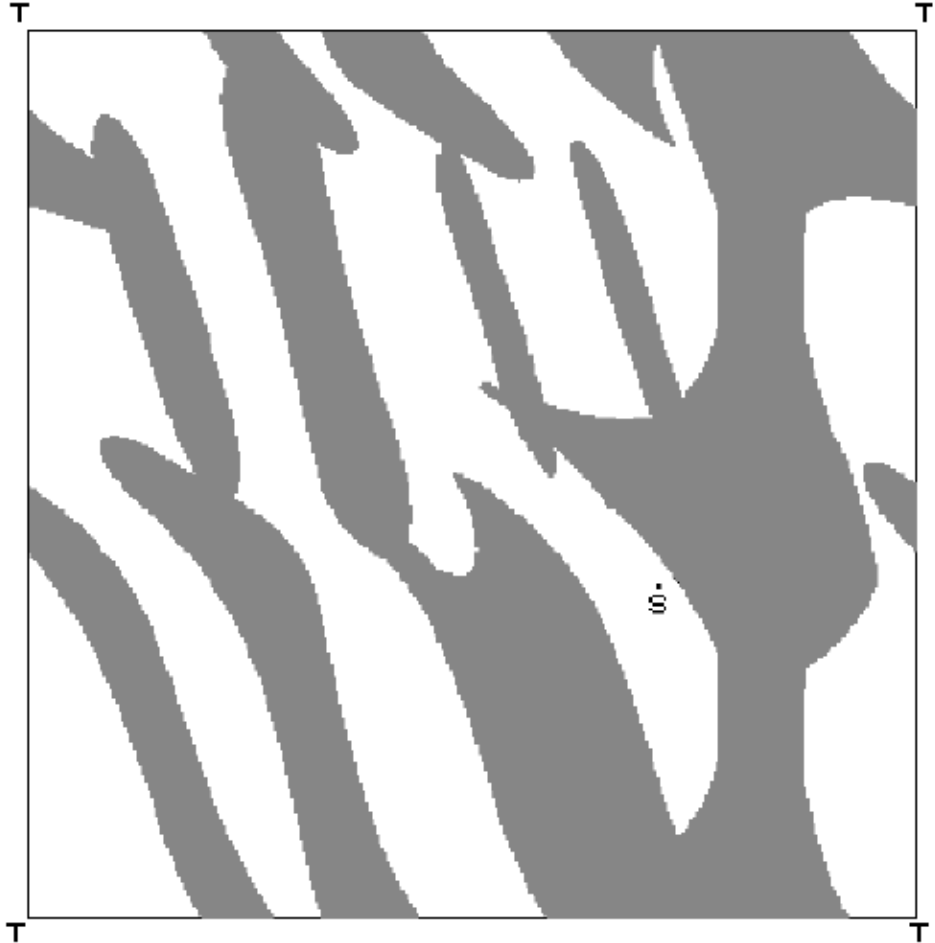,width=5in} 
\end{tabular}
} 
%\figurecaption{ 
\caption{\small The sample task of Fig. ~\ref{task}; C-space representation. }
\label{taskcspace}
\end{figure}

Figure ~\ref{taskcspace} gives the C-space representation of W-space of Figure
~\ref{task}. Angle $\theta_1$ is along the horizontal axis, $\theta_2$ - 
along the vertical axis. The range of change of each angle is $2\pi$, making
C-space a square. The dark areas represent the virtual obstacles.
The C-space correspondence to a two-torus means in that all four
corners of the square are identified (i.e. correspond to the same
point). Similarly, the top and bottom edges of the square are
identified, and so are the left and right edges. Given this last
fact, note that the C-space in Figure ~\ref{taskcspace} contains only 
one virtual obstacle which corresponds to four physical obstacles in W-space,
Figure ~\ref{task}. Point T is chosen as the corner point of the C-space
square; in principle, therefore, one's moving from point S to any
corner will produce a legitimate (if not necessarily the shortest)
path for the arm in W-space.

\subsection{Characteristics of the Configuration Space Control}

This mode of control has several distinct advantages (see also
Results, Section ~\ref{sec:Results}):

\begin{itemize}
\item From the operator standpoint, the task is simplified greatly:
instead of dealing with a complex jointed kinematic structure, the
operator has to solve a simple maze-searching problem with complete
information, which humans are very good at.

\item One explanation for the task simplification is that the
responsibilities are divided in this mode - the operator can think
of the motion planning only, while the computer takes on the
problem of collision analysis.

\item The arm's actual motion is quickly and easily calculable from user
input, guaranteeing good real-time performance.

\item Unlike in W-space, performance here does not seem to depend much on
the obstacle layout. Indeed, this mode has consistently yielded near
optimal performances by the human operator in a variety of
settings. This is consistent with the fact that humans can easily
``see'' the path in a bird-eye view of a fairly complex maze, while
they have difficulty visualizing a path in a simple scene with an
arm manipulator (see Figure ~\ref{task}).  The operator easily discards many
``dead-end'' directions in the maze representation, but find it
difficult to identify them in Figure ~\ref{task}. 

\item The mode requires very little training, mostly to get used to the
peculiarities of flat presentation of two-torus - e.g. to the fact
that once the point reaches the top edge of the C-space square, it
appears at the bottom edge. In fact, performance has been just as
good for an inexperienced user as for an experienced one.

\item Unlike the W-space control, the subject can often easily see if a
solution (a path) exists. In fact, it is this kind of
decision-making that the operator uses extensively along the way to
discard potential dead-ends.
\end{itemize}

A few drawbacks deserve to be noted of this mode, although
their impact is not nearly as great as those in W-space control:

\begin{itemize}
\item The fact of dealing with an abstract (C-) rather than physical (W-)
space may make it difficult for the operator to address some global
navigation tasks, such as choosing targets for the arm to
reach. This problem is easily avoided if the corresponding W-space
view is drawn in parallel with the C-space used by the operator (see
Figures ~\ref{task} and ~\ref{taskcspace}).

\item While extremely helpful in 2D, the mode is not likely to easily
generalize to more complex multi-link systems see discussion in
Section ~\ref{sec:Discussion}).

\item Computation of C-space is an expensive operation which must be performed
to satisfy the complete information model (see Section ~\ref{sec:Discussion} 
for details on the proposed uncertainty model).
\end{itemize}

\section{The Interface}
\label{sec:Interface}

The current version of the user interface has several interesting
features:

\begin{itemize}
\item The user can generate - e.g., for practice - custom or random
obstacle environments around the arm.

\item Both C-space and W-space displays are provided. As mentioned above,
the W-space window is a good tool for visualizing global navigation
tasks, such as deciding on a target position for the arm
endpoint. The user can, for example, define the target in terms of
the arm endpoint Cartesian coordinates $(x, y)$, and the computer will
recover the corresponding arm configuration through inverse
kinematic equations (1, 2).

\item Motion control in C-space is done via mouse interaction; joystick
control is being considered, especially in the future 3D extension.

\item For experimental purposes, the user can switch back and forth
between C-space and W-space control.

\item The simulation keeps track of the time elapsed and path traversed, to
help compare the two control methods.
\end{itemize}

\section{Results and Discussion}
\label{sec:Results}

\subsection{Results}

Overall, the proposed C-space control mode performed admirably when
compared to the traditional W-space control. Current results
(achieved with a C-space interface version that is still under
development) show improvement in performance on the order of
magnitude when switching from W-space control to the proposed
C-space control. The path produced approaches the optimal (shortest)
path and time to complete the task.  Further, the cognitive part of the time 
spent in the case considered is negligible, since the 2D mazes produced by the
virtual obstacles are simple to navigate and to learn. This
remarkable fact puts the human operator ahead of the existing
computer algorithms, contrary to the W-space control where human
performance has been much worse. It also suggests interesting
questions and extensions to more difficult 3D cases.

Table 1 summarizes information from a series of controlled
experiments performed in 1996-97 at the UW Robotics Lab, to test
human performance in motion planning tasks.  One of the tasks given
to the human subjects was to move a two-link arm, very similar to the
one considered in this paper, from the start to target
configurations. Only W-space control was available (Section ~\ref{sec:Wspace}). 
In the table, the path length is the integral of both joint angle
changes during the motion; also given is the time ((in seconds)
taken by subjects to complete the task. The data given represents
the performance of 12 subjects on the second day of tests, after
training and practice on the previous day. (The results on
48 untrained subjects, in tests with a simulated as well as physical
arm manipulator, were quite similar).  A full analysis of this work
can be found in \cite{Liu:b}.

\begin{table}[htb]
\caption{Descriptive Statistics}
\centering{
\begin{tabular}{|c|c|c|c|c|} \hline
Variable & Mean & Minimum & Maximum & Stand. Dev \\ \hline\hline
path length & 129.04 & 15.13 & 393.90 & 107.99 \\ \hline
time & 504.83 & 90.00 & 900.00 & 365.89 \\ \hline
\end{tabular}
}
\end{table}

\begin{figure}[t]
\centerline{
\begin{tabular}{c}
\psfig{figure=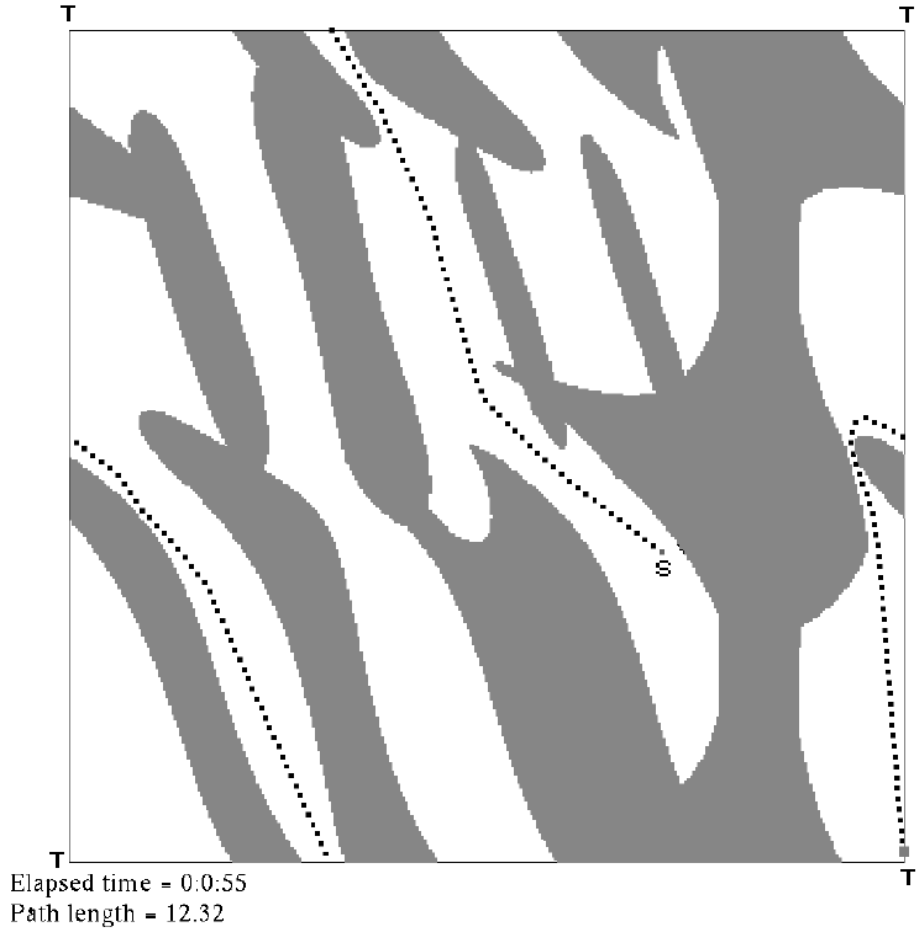,width=5in} 
\end{tabular}
} 
%\figurecaption{ 
\caption{\small The sample task of Fig. ~\ref{task}: C-space motion control.
The corresponding W-space in Fig. ~\ref{wspacetaskdone}. }
\label{taskcspacedone}
\end{figure}

No similar study was carried out for the C-space control mode.
However, based on the observations and tests by these authors, the
study is not necessary: the performance improvement is very clear
and consistent. Further, it is clear that in the task of Figure ~\ref{task}
different subjects are likely to produce almost the same (nearly
optimal) path, with the mean path length of about 12, the standard
deviation of about zero, and the mean time below 1 min.  The
path length and time values in Table 2 show an order of magnitude
improvement compared to the data on W-space control in Table
1. Sample results from 5 consecutive runs of C-space control are
given in Table 2.  One of these runs is shown in Figures ~\ref{taskcspace} 
and ~\ref{taskcspacedone}.

\begin{table}[htb]
\caption{Sample Runs}
\centering{
\begin{tabular}{|c|c|c|c|c|c|} \hline
Variable & Run 1 & Run 2 & Run 3 & Run 4 & Run 5 \\ \hline\hline
path length & 12.67 & 12.39 & 12.24 & 12.27 & 12.28 \\ \hline
time & 56 & 54 & 53 & 53 & 54 \\ \hline
\end{tabular}
}
\end{table}

\begin{figure}[t]
\centerline{
\begin{tabular}{c}
\psfig{figure=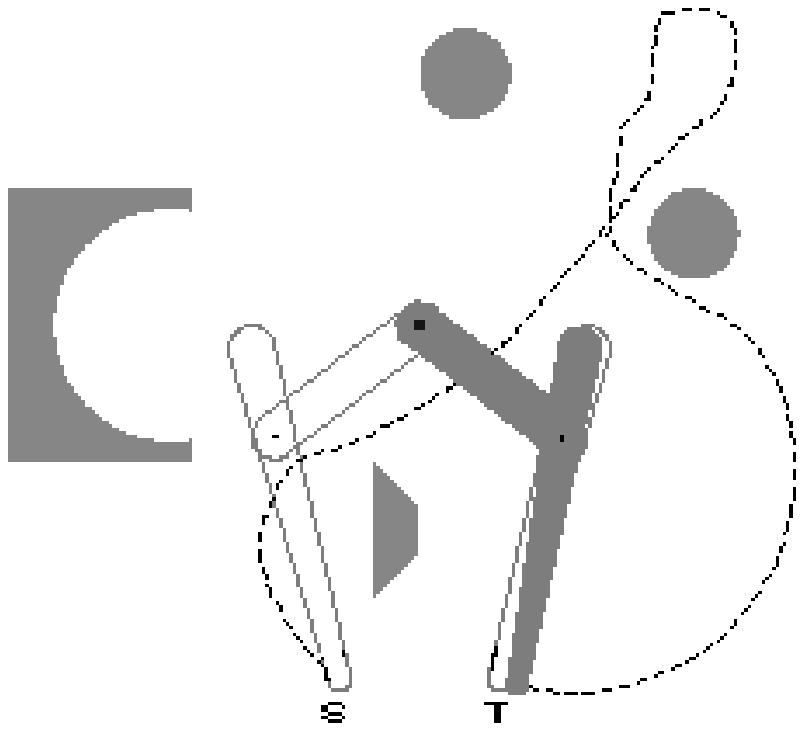,width=5in} 
\end{tabular}
} 
%\figurecaption{ 
\caption{\small The W-space view of the task in Fig. ~\ref{taskcspacedone}. 
The path produced does not contain unnecessary ``detours'' common to 
W-space control (see Fig ~\ref{examp1}), and approaches optimal path 
for this task.}
\label{wspacetaskdone}
\end{figure}

The consistency between these runs - both in path length and
completion time - is very similar to the subjects performance in a
common maze-searching problem. It also stands in contrast to the wide
range of results produced in the W-space model.  This suggests that the 
proposed transformation to C-space control does indeed make the task at hand
similar to the maze-searching task.

\subsection{Discussion}
\label{sec:Discussion}

This paper proposes an approach to human-guided teleoperation of a
robot arm manipulator based on the configuration space (C-space)
rather than on the common work space (W-space) control. Instead of
directly confronting the problem of collision analysis, which is
known to be extremely challenging for the human spatial reasoning,
the task is offered to the operator in C-space where one can
concentrate on global navigation, leaving collision analysis to the
computer. Thus reduced task becomes a maze-searching problem in
which humans are known to be very good. Designing such a system
takes, first, calculation of the C-space, and second, an adequate
user interface.

While this approach can be immediately useful even in its
two-dimensional version described, in order to become a truly
universal tool it needs to be extended to the three-dimensional case
and to more degrees of freedom.  The advantage for the operator of
dealing with a point rather than a complex jointed kinematic
structure is obvious. The challenge is to produce an adequate user
interface (specifically, develop ways of visualizing and guiding a
point in a higher-dimensional space) and to do C-space calculation
and collision analysis fast enough to keep the operator active at
the control station. One possibility here is to help the operator
handle the environment with incomplete, rather than complete,
information; this would mean a significant reduction in the C-space
computation costs. Success in this area will also mean applicability
of the approach to a dynamic environment with moving obstacles.
Computer algorithms for motion planning with incomplete information
are available (e.g. \cite{Lum}).  Experiments with human subjects operating
in an unknown maze \cite{Liu:a, Liu:b} suggest that humans might be able to
handle this case as well.

The immediate problem is to determine if the resulting three dimensional
C-space will still be as helpful to the human in performing the task as the 
two dimensional case was.  Also necessary will be: algorithms for
computing C-space for a multi-link arm; algorithms for collision
analysis in 3D space; procedures for C-space visualization, and for
arm motion control in W-space. These are likely to raise issues of
computational complexity and real-time control.

\end{document}